**VESTIGE: A Knowledge-Guided Masking Strategy for Corruption-Aware Fine-Tuning of Genomic Transformers, Validated on Ancient DNA Reconstruction**

Angshuman Chakravertty[a], Rahul Maheshwari[b,*]

[a]School of Technology Management & Engineering, SVKM's Narsee Monjee Institute of Management Studies (NMIMS) Deemed-to-be-University, Jadcherla, Hyderabad 509301, Telangana, India, Hyderabad, India

[b]School of Pharmacy and Technology Management, SVKM's Narsee Monjee Institute of Management Studies (NMIMS) Deemed-to-be-University, Hyderabad, Jadcheral-509301, Telangana, India

***Address for correspondence***

Dr. Rahul Maheshwari

School of Pharmacy and Technology Management,

*SVKM's Narsee Monjee Institute of Management Studies (NMIMS) Deemed-to-University, Hyderabad, Jadcheral*-509301, Telangana, India

E-mail: rahulgs.maheshwari@gmail.com and rahul.maheshwari@nmims.edu

Tel/Fax: +91 9893338868

*Disclosure: There is no conflict associated with this manuscript*

**Abstract**

Standard masked-language-model fine-tuning applies a uniform masking probability across every token position, implicitly assuming that reconstruction difficulty is position-agnostic. When the degradation process is known, characterised, and concentrated at predictable positions, this assumption fails: at peak damage sites the model can underperform a frequency-matched random predictor. We introduce VESTIGE, a system with a parameter-free, drop-in replacement for the standard MLM collator that aligns the self-supervised masking distribution with an empirically measured per-position corruption profile. The collator requires no architectural change, no additional parameters, and no modification to the fine-tuning objective - only that a degradation profile be supplied. As a validating instance, we apply VESTIGE to ancient DNA (aDNA) reconstruction, where cytosine deamination produces a sharply position-dependent C→T (5') and G→A (3') gradient that decays exponentially into the fragment interior and is quantified at per-position resolution by mapDamage2. Spatial redistribution is isolated as the sole variable by rescaling the position-weighted distribution so that the mean masking rate over C/G positions equals 15% - identical to standard MLM - with model, data, seed, and hyperparameters held fixed across both fine-tuning runs on DNABERT-2 with a mammoth CDS corpus (two specimens; three training and four held-out genes). Across six terminal-zone widths in {3,5,10,15,20,25} - spanning inner-terminal to moderately wide zones - and 626 paired evaluation windows, VESTIGE leads standard MLM at every width ($\Delta$=+4.18 to +10.35 pp, all $p \lt {10}^{-8}$), reduces validation cross-entropy by 13% (3.274 vs. 3.757), and produces ESMFold reconstructions with TM-score >0.95 across all six reconstructions (three genes) even under damage amplified 10–30× beyond authentic PMD rates. A 1D CNN biosecurity classifier returns AUC =0.935 on a held-out virulence-gene test set and clears 98.2% of reconstructed windows, with the remaining 1.76% attributable to reference-genome features rather than reconstruction artefacts. The principle is domain-agnostic: any degradation profile with a measurable position- or context-specific frequency - FFPE artefacts, bisulfite-conversion gradients, low-coverage or metagenomic noise, nanopore homopolymer errors - substitutes directly for the PMD array, making VESTIGE a knowledge-guided training routine for intelligent systems that operate on degraded or noisy sequence inputs.

## 1. Introduction

Self-supervised masked language modelling is the de facto pre-training and fine-tuning objective of every major transformer-based genomic language model (Dalla-Torre et al., 2024; Ji et al., 2021; Nguyen et al., 2023; Zhou et al., 2023). Its default collator applies a uniform 15% masking probability across every token position (Devlin et al., 2019) - a position-agnostic assumption that holds only when reconstruction difficulty is itself position-agnostic. When the difficulty is concentrated - when, that is, the corruption process is characterised, predictable, and operant at a specific subset of positions - uniform masking is not merely suboptimal; it is actively harmful. Drawing more training signal from positions that are already easy, while leaving difficult positions under-trained, produces a representation tuned to the wrong objective. We argue that whenever an empirically measured corruption profile is available, the masking distribution of a transformer fine-tune should reflect the empirical concentration of the corruption - a principle we call *corruption-aware fine-tuning*. The present paper validates this principle through *VESTIGE*, a parameter-free, drop-in replacement for the standard data collator that realigns the masking distribution with an externally measured degradation profile.

As a validating instance, VESTIGE targets ancient DNA (aDNA) reconstruction, where cytosine deamination produces a sharply position-dependent C→T (5') and G→A (3') substitution gradient that decays exponentially into the fragment interior (Shapiro & Hofreiter, 2014). Tools such as mapDamage2 (Ginolhac et al., 2011; Jónsson et al., 2013) quantify this gradient at per-position resolution, yet standard analysis pipelines either discard the profiled frequencies after quality control or repurpose them only to discard damaged reads - pmd-mask (Lefeuvre, 2023) hard-masks PMD sites to N for variant calling, discarding the signal rather than exploiting it. VESTIGE closes this gap by feeding the same mapDamage2 output directly into a damage-aware masking collator: the probability for each C position is proportional to the empirical C→T frequency at that position from the fragment 5' terminus, and symmetrically for G positions from the 3' terminus, with A/T positions receiving zero masking probability because deamination is C/G-specific. The collator requires no architectural change, no additional parameters, and no modification to the fine-tuning objective; the empirical distribution is rescaled so that the mean masking rate over C and G positions equals 15% - identical to standard MLM (Wettig et al., 2023) - isolating spatial redistribution as the sole experimental variable.

The principle that the masking schedule shapes the learned representation is well established in NLP - dynamic masking regenerates the pattern per epoch (Liu et al., 2019), span masking forces recovery of contiguous dependencies (Joshi et al., 2020), and adaptive PMI-Masking selects tokens by content or difficulty (Levine et al., 2020), with curriculum strategies (Bengio et al., 2009) providing further schedule structure. Tang et al. (Tang et al., 2026) recently showed that span and blockwise masking extract high-level features, where random masking captures only low-level statistical texture. Yet every existing DNA LM, despite inheriting from this lineage, retains the BERT-original uniform-random distribution; none injects an externally measured physical corruption prior. VESTIGE is the first to do so, deriving the masking distribution from empirical biological degradation rates rather than from randomness, contiguity, or learned difficulty. This positions DAM as a general intelligent-systems training routine for any sequential setting in which corruption concentrates at measurable, predictable positions - a category substantially broader than ancient DNA, as developed in Section 6.

The end-to-end VESTIGE pipeline integrates three validation layers that no prior reconstruction work has combined. (i) Reconstruction: DNABERT-2 (Zhou et al., 2023) with BPE tokenisation (Sennrich et al., 2016) is fine-tuned on a woolly mammoth CDS corpus drawn from two specimens (Lynch et al., 2015; Palkopoulou et al., 2015) spanning approximately 4,300 to approximately 44,800 years BP. (ii) Structural plausibility: ESMFold (Lin et al., 2023) - the single-sequence language-model protein-structure predictor - receives

DAM-reconstructed CDS, and the predicted folds are superposed against the Asian elephant reference (Panis et al., 2026) using Kabsch rotation (Kabsch, 1976) and scored by TM-score (Zhang & Skolnick, 2004) and Cα-RMSD. (iii) Biosecurity screening (Diggans & LeProust, 2019) using a 1D CNN (Park et al., 2023) trained across 11 pathogen virulence-gene classes flags any reconstructed window whose maximum class probability exceeds a fixed decision threshold. The controlled ablation isolates the masking distribution as the sole variable by holding model, data, seed, and hyperparameters fixed across DAM and MLM runs (matched at 15% C/G density), so that any DAM–MLM difference is attributable purely to spatial redistribution. The headline empirical finding is that at the innermost terminal-zone width (T_END = 3, the peak PMD positions), standard MLM fine-tuning recovers only 20.49% of damaged nucleotides - below the 27.67% achievable by a frequency-matched random predictor - whereas DAM achieves 30.84% ($p = 1.33 \times 10^{-10}$, $n = 486$). DAM leads MLM at every one of six terminal-zone widths ($\Delta = +4.18$ to $+10.35$ pp, all $p < 10^{-8}$, $n = 626$ paired windows), produces 13% lower validation cross-entropy (3.274 vs. 3.757), and yields ESMFold reconstructions with TM-score > 0.95 across 3 training genes even under damage amplified 10–30× beyond authentic PMD rates. The biosecurity layer clears 98.2% of reconstructed windows with AUC = 0.935 on a held-out virulence-gene validation set, with all 11 flagged windows attributable to reference-genome features (conserved ankyrin repeats, high-GC compositional bias) rather than to reconstruction artefacts.

In this paper, we first review related work spanning aDNA damage modelling, DNA language models, the evolution of masking strategies, and downstream validation. Next, we provide the materials and methods, including the controlled ablation design, the DAM collator specification, the in silico damage simulation, and the evaluation protocol. Then the results across validation loss, nucleotide recovery (background and terminal), terminal-zone sensitivity, structural validation, and biosecurity screening are presented. We thoroughly discuss the generalisability to FFPE (Do & Dobrovic, 2014), bisulfite-conversion, low-coverage and metagenomic (Quince et al., 2017), and nanopore homopolymer-error (Rang et al., 2018) settings, together with limitations and future directions including curriculum masking (Bengio et al., 2009) and multi-reference alignment (Günther & Nettelblad, 2019).

## 2. Related Work

Ancient DNA fragments are characteristically short, around 30-70 bp, with the cytosine deamination producing C→T transitions at the 5' termini and G→A transitions at the 3' termini, as can be understood from (Orlando et al., 2021). mapDamage, the tool being used, is the standard damage profile tool (Ginolhac et al., 2011; Jónsson et al., 2013), along with PMDtools, as it complements it by authenticating the endogenous reads using the damage signal (Skoglund et al., 2014). Importantly, these tools tend to characterise damage but remain decoupled from any kind of downstream model and the empirical damage frequencies that they produce get discarded after quality control and are never repurposed to guide the sequence reconstruction. Importantly, pmd-mask (Lefeuvre, 2023) establishes it as prior art, where it consumes mapDamage output and hard-masks PMD sites to N, where the purpose is variant calling; it discards the damage signal, whereas VESTIGE uses the same damage signal as a masking input rather than discarding it. VESTIGE closes this gap by feeding the profiled frequencies directly into a damage-aware masking collator.

The first model of its kind to be used with DNA was the BERT-style DNABERT model (Ji et al., 2021) with the help of 6-mer tokenisation and established it as the baseline masked LM paradigm for the genomic sequence. Nucleotide Transformer (Dalla-Torre et al., 2024) and HyenaDNA (Nguyen et al., 2023) have extended this paradigm to larger parameter counts and longer sequence contexts, and both retain the uniform random masking, which is identical to the standard MLM. DNABERT-2 (Zhou et al., 2023) with BPE tokenisation and multi-species pre-training is the fine-tuned backbone in VESTIGE, but the shared limitation across all these models is that they apply position-agnostic masking, where the masking probability is uniform across all the nucleotide positions regardless of their biological damage status, and none of them exploit the empirical damage gradients, and DAM addresses this gap.

The standard BERT (Devlin et al., 2019) uses a uniform random masking of 15%, making it the baseline from which VESTIGE departs. Furthermore, dynamic masking introduced by RoBERTa (Liu et al., 2019) regenerates the masking pattern in every epoch instead of fixing it at preprocessing, which shows how the masking schedule affects what is learned, but the distribution stays uniform in expectation, and span masking, as in SpanBERT (Joshi et al., 2020), demonstrated how masking contiguous token spans instead of random individual tokens changes what the model learns; the span masking forces the model to recover contextual dependencies rather than local texture. Adaptive masking such as PMI-Masking (Levine et al., 2020), chooses what to mask by difficulty or content rather than uniformly, and Tang et al. (Tang et al., 2026) extended this finding to masking image modelling, where random patch masking captures only low-level features, while at the same time, blockwise and span masking elicit high-level contextual representations and thus directly paralleling the DAM motivation. From all this, the general principle formulated is that if a masking distribution shapes the learned representation, then a masking distribution extracted from empirical biological damage rates should actually produce representations that are maximally informative for the damage-specific reconstruction, and DAM operationalises this for aDNA.

All the masking techniques such as dynamic, span, and adaptive derive the masking decision from within the model/data, such as randomisation schedule, a contiguity heuristic, or learned difficulty, but DAM derives the masking distribution from an external, measured corruption prior (mapDamage2 PMD frequencies) and not from the model, text content, or any fixed heuristic. DAM is the first masking strategy to connect the masking distribution with an empirically characterised physical degradation process, making it transferable to any domain with a known corruption profile.

AlphaFold2 (Jumper et al., 2021) demonstrated near-experimental accuracy for protein structure prediction from sequence alone. Additionally, ESMFold (Lin et al., 2023) extends this to a single language-model inference pass without the alignment of multiple sequences, and this is why ESMFold was chosen for validating the reconstructed mammoth CDS sequences in Section 4.4. It is important to note that the structural validation of computationally reconstructed sequences is an established step in paleoproteomics pipelines, and this ensures that the reconstructed CDS encodes a plausible fold. For biosecurity, the dual-use risk of sequence reconstruction tools, where the same computational pipeline that recovers the ancient mammoth CDS could reconstruct the pathogen sequences, was identified as a cybersecurity concern in (Peccoud et al., 2017), and it is also a standard practice in DNA synthesis pipelines to screen reconstructed windows against the known virulence gene databases (Diggans & LeProust, 2019), which is further motivated by Park et al.'s (Park et al., 2023) demonstration that robust out-of-distribution detection is essential when classifying sequences unseen during training. All three validation layers (sequence recovery, structural plausibility, and biosecurity screening) are integrated by VESTIGE as an end-to-end pipeline, which distinguishes it from any kind of prior DNA language model work that evaluates only the reconstruction accuracy.

## 3. Materials and Methods

### 3.1 Dataset and Specimens

For the specimen choice the woolly mammoth (*Mammuthus primigenius*) was selected as a model system for aDNA reconstruction primarily for its well-characterised post-mortem damage profiles and the high availability of public sequencing data. Table 1 below shows the specimen details.

Table 1. Specimen details from project ERP008929.

| Specimen | Region | Before Present (BP) |
|---|---|---|
| ERR855944 | Wrangel Island | ~4,300 |
| ERR852028 | Oimyakon, Siberia | ~44,800 |

Both of these specimens are from Sequence Read Archive (SRA) project ERP008929 (Palkopoulou et al., 2015).

For the purpose of aligning targets and CDS sources, the genome *Elephas maximus* reference GCF_024166365.1 (EanMak 1.0 / mEleMax1 primary haplotype) was selected (Panis et al., 2026). For fine-tuning and evaluation, three protein-coding genes were selected, namely TRPV3, KCNK9, and HBB (annotated as LOC126080006 in EanMak 1.0), all of which are functionally relevant to cold-adaptation in proboscideans(Lynch et al., 2015); the end-to-end pipeline is summarised in Figure 1.

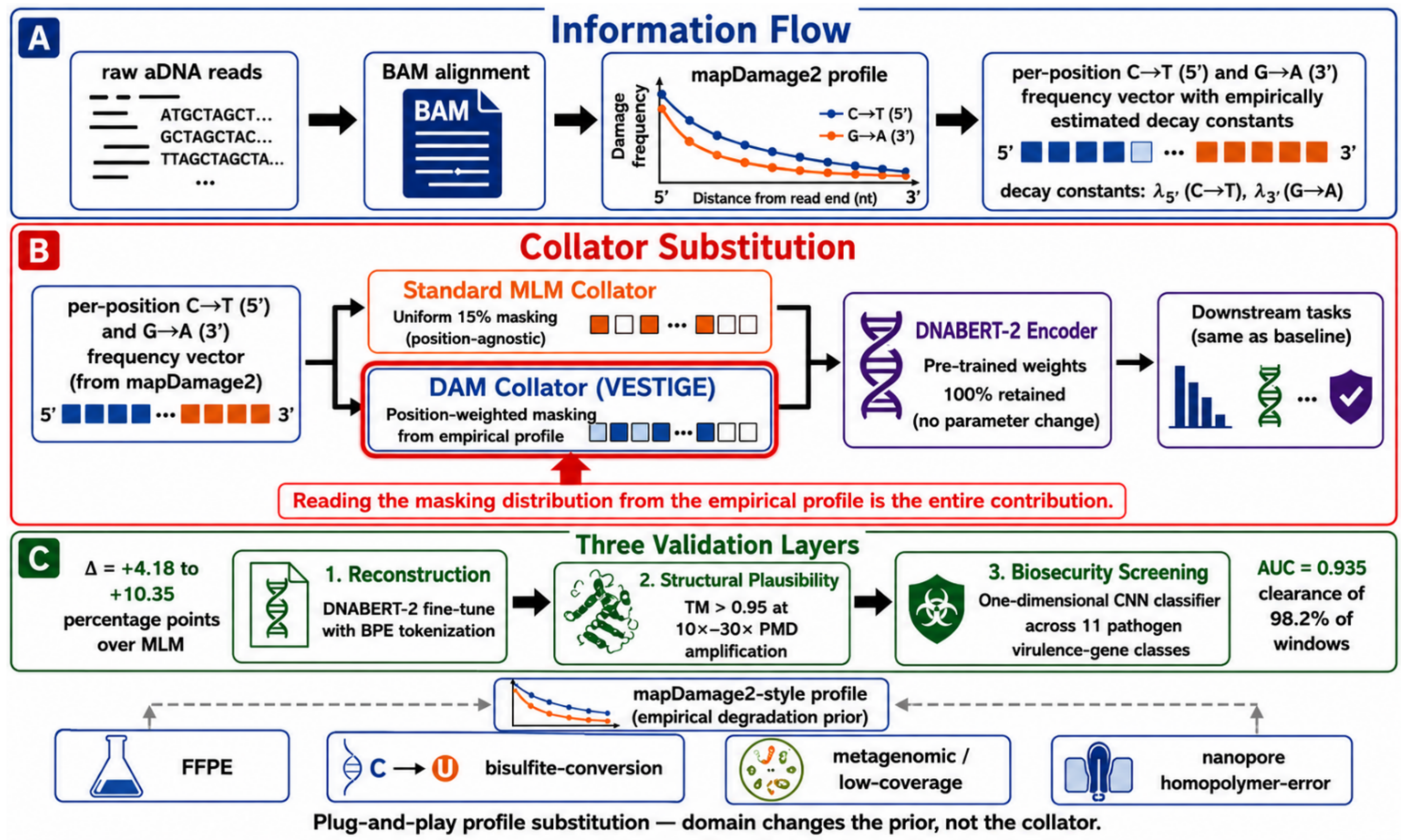

**Figure 1. The VESTIGE end-to-end pipeline - the single methodological substitution is the masking collator at the centre, and everything upstream and downstream of it is unchanged.** The figure is structured as three horizontal tiers plus a domain-extensibility strip. **Top tier - information flow:** raw ancient-DNA reads from NCBI SRA accessions ERR855944 and ERR852028 (Palkopoulou et al., 2015) are aligned to the *Elephas maximus* reference genome (GCF_024166365.1, EanMak 1.0) (Panis et al., 2026) using aDNA-tuned BWA-aln parameters (-l 16500, -n 0.01) (Li & Durbin, 2009), and per-position 5'-C→T and 3'-G→A substitution frequencies are quantified by mapDamage2 (Ginolhac et al., 2011; Jónsson et al., 2013). The two mammoth specimens yield peak C→T frequencies of 0.29% and 0.41% at position 1, decaying to a shared ~0.1% background by position 10. **Middle tier - collator substitution (visually dominant, ringed in red):** the per-position damage profile vector enters two parallel collators - the standard MLM collator (Devlin et al., 2019) applying uniform 15% masking (drawn in MLM-orange) and the VESTIGE damage-aware collator applying position-weighted masking proportional to empirical 5'-C and 3'-G deamination frequency (drawn in DAM-blue). Both collators feed the pre-trained DNABERT-2 encoder (Zhou et al., 2023) with **100% of pre-trained weights retained, zero parameter change, zero architectural modification**, isolating spatial mask redistribution as the sole experimental variable at matched 15% mean C/G density (rescaling given by Equation [eq:scale]). **Bottom tier - three validation layers integrated in one pipeline:** (i) sequence reconstruction via DNABERT-2 with BPE tokenisation (Sennrich et al., 2016; Zhou et al., 2023), where DAM beats MLM by $\Delta = +4.18$ to $+10.35$ percentage points across six terminal-zone widths (all $p < 10^{-8}$, $n = 626$ paired evaluation windows); (ii) structural plausibility via ESMFold (Lin et al., 2023) superposed against the *Elephas maximus* reference fold (Panis et al., 2026) using Kabsch rotation (Kabsch, 1976), yielding TM-score > 0.95 across all six reconstructions (three genes) even under PMD amplified 10×–30× beyond empirical rates; (iii) biosecurity screening via 1D CNN classifier (Park et al., 2023) over 11 pathogen virulence-gene classes, clearing 98.2% of reconstructed windows with AUC = 0.935 on a held-out validation set, every flag attributable to reference-genome features (ankyrin repeats in TRPA1, high-GC composition in FASN) rather than reconstruction artefacts (Diggans & LeProust, 2019; Peccoud et al., 2017). **Domain-extensibility strip:** four plug-and-play analogues - FFPE tumour-specimen damage (Do & Dobrovic, 2014), bisulfite-conversion data, low-coverage and shotgun metagenomic noise (Quince et al., 2017), and nanopore homopolymer-error profiles (Rang et al., 2018) - share the same DAM collator interface; only the empirical prior vector changes, not the encoder.

The corpus is deliberately compact with just 2 specimens, 1 reference genome, 3 genes for fine-tuning (344 windows, 275 train/69 val), and 4 more genes held out for evaluation. This supports a controlled ablation with DAM against MLM, identical model, data, and seed, differing only in the masking distribution. The effect that is being tested is a within-pair contrast; thus, the statistical power does not come from training-corpus size but rather from the 626 evaluation windows and per-window pairing, and a proof-of-concept does not require scale. The 4 held-out genes (TRPA1, UCP1, ADRB3, FASN) were absent from fine-tuning, but DAM showed the same effect for them too, and thus the results were not memorised from the three training genes. The design scales trivially with more specimens or species, adding only windows while leaving the collator unchanged, and broader multi-species validation is discussed further in Section 6.

### 3.2 Alignment and Damage Profiling

Once the data was acquired, the raw FASTQ reads were subsampled to 5 million reads per sample using a command-line toolkit called seqtk (seed 42) (Li, 2013). The primary issue with aDNA is that the reads are characteristically short, around 30-70 bp and also exhibit elevated mismatch rates due to post-mortem deamination; using the standard BWA-MEM2 with default seed length is inappropriate for this distribution(Schubert et al., 2012). The BWA-aln was then used with two aDNA-tuned parameters, -l 16500 which disables seeding by setting the seed length above the realistic read length and also forces the full-read alignment, -n 0.01 which allows a mismatch of up to 1% to tailor deamination-induced substitution(Li & Durbin, 2009).

The data was sorted and the indexed BAM files produced via bwa samse were piped to samtools sort, where the specimen ERR852028 mapped 4,966,693/5,000,000 (99.33%) and ERR855944 mapped 4,956,238/5,000,000 (99.12%). The mapping rates greater than 99% confirm adequate sample quality for downstream damage profiling.

Subsequently, mapDamage2 is implemented for each sorted BAM in comparison to the reference FASTA which produces per-position C→T (5' terminus) and G→A (3' terminus) substitution frequency tables from the misincorporation.txt output(Jónsson et al. 2013)(Ginolhac et al. 2011). It was observed that the C→T frequency at position 1 was 0.0029 for ERR855944 and 0.0041 for ERR852028 decaying to a shared background of ~0.001 by position 10, which indicates the exponential terminal decay of the canonical aDNA deamination signature (Briggs et al., 2007). It is noteworthy that the mean of both specimen profiles (ct_5p and ga_3p arrays with 70 positions each) was stored as a numpy array for use by the DAM collator, and the higher terminal damage in the older ERR852028 relative to ERR855944 is consistent with the known age-damage relationship in aDNA (Dabney et al., 2013). Figure 2 shows the resulting C→T and G→A frequency profiles for both the specimens.

It should be emphasised that mapDamage2's built-in R plotting was bypassed due to R ≥ 4.x incompatibility and to visualise the damage profiles custom Python scripts were used.

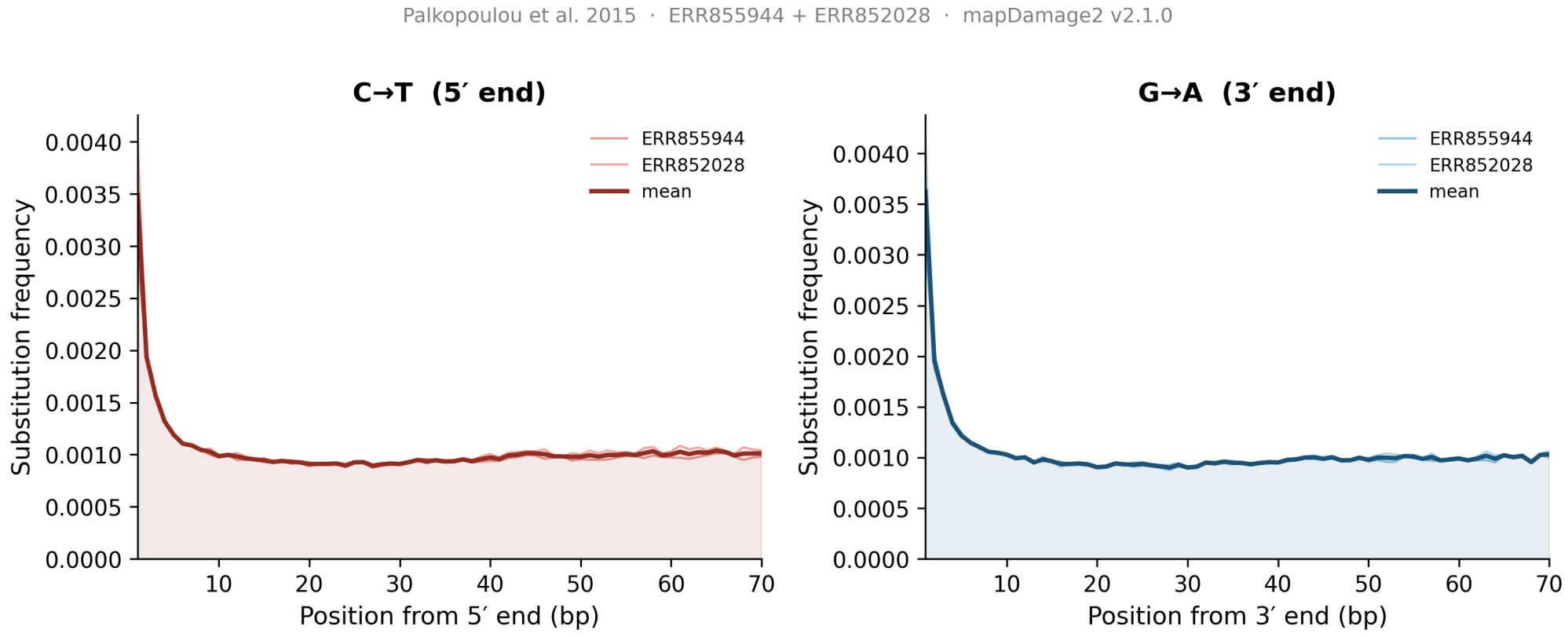

***Figure 2.*** *Post-mortem damage profiles (C→T and G→A) for both woolly mammoth specimens by read-end position.*

### 3.3 Damage-Aware Masking Collator

In HuggingFace-style transformer fine-tuning, the data collator is the small piece of training code that decides which tokens to mask on each forward pass before the model is asked to recover them. The default collator uniformly selects masks with a fixed 15% probability per token. VESTIGE replaces only this collator: instead of masking a fixed 15% of every token, it masks a position more often if that position is empirically known to be corrupted, and a position less often (down to zero) if it is not. The model, the tokenisation, the optimiser, the loss function, and the weights are unchanged. The single substitution is the masking distribution itself, derived from an externally measured per-position corruption profile fed in at collator construction time.

The standard DataCollatorForLanguageModeling applies uniform masking probability of 15% across all the token positions without consideration for a token's semantic identity or its relative position within the sequence fragment (Devlin et al., 2019), which results in it treating all positions equally and disregarding the position-specific damage gradient in the aDNA, where C→T substitutions are strongly concentrated at the 5' fragment termini. Therefore, a model trained this way never learns to develop specialised reconstruction ability at the exact position where the ancient sequence information is most degraded.

As illustrated in Figure 1, DamageAwareDataCollator then utilises the empirical mapDamage2 PMD profiles to override the masking probability matrix by assigning per-position, per-base masking probabilities. This makes it so that the masking gradient mirrors the biological damage gradient, where the positions with higher empirical damage would receive proportionally higher masking probability during training.

$$P_{\text{mask}}(p,b) = \begin{cases} \text{ct}_{5p}[\min(p,\ 69)] & \text{if } b = \text{C} \\ \text{ga}_{3p}[\min(L-1-p,\ 69)] & \text{if } b = \text{G} \\ 0 & \text{otherwise} \end{cases}$$

As defined in Equation [eq:dam], the $\text{ct}_{5p}$ and $\text{ga}_{3p}$ are the means of C→T and G→A frequency arrays, respectively, derived from damage_profile.npy which consists of 70 positions and is averaged across both specimens; L represents the sequence length; b is the first nucleotide of the BPE token at position p, and A/T tokens are the ones that receive zero masking probability, which is biologically consistent since deamination is C/G-specific.

There persisted an issue where the empirical damage frequencies peaked at 0.29-0.41% at position 1, which was far below the 15% MLM masking rate. Applying the raw frequencies would result in DAM ~40x fewer masked tokens per batch compared to MLM, making any loss comparison invalid. To resolve this the probability matrix is uniformly rescaled so that the mean masking rate over C/G positions equals 15%.

$$P'_{\text{mask}}(p,b) = P_{\text{mask}}(p,b) \cdot \frac{0.15}{\bar{P}_{\text{CG}}}$$

As defined in the Equation [eq:scale], $\bar{P}_{\text{CG}}$ is the mean of $P_{mask}$ over the C and G positions only, which ensures that the ablation is controlled and identical to the total masking density on the C/G positions between DAM and MLM where they differ only in the spatial distribution. After scaling, the terminal C at position 1 receives approximately 28% masking probability, the interior C at position ~130 receives approximately 14.8%, and MLM applies a flat 15% to all the positions. The standard BERT follows an 80/10/10 replacement schedule where 80% gets masked with [MASK], 10% have random tokens, and 10% remain unchanged and are then retained identically for both collators (Devlin et al., 2019).

To verify the correct implementation of the DAM collator, a chi-squared test of independence was conducted on masking position counts across 10,000 simulation trials. The difference between the terminal positions 1-5 from the fragment end and the central positions 21-50 is $\chi^2(1) = 7154.8, p < 10^{-3}$, which confirms that, as intended, the masking is significantly concentrated at terminal positions.

The collator shows dependence only on the damage array as the input, so even if the array is swapped, the code remains unchanged. FFPE-fixed DNA with its own deamination signature, bisulfite sequencing with its C→T conversion gradient, and RNA degradation profiles are clear biological analogues, and platform-specific sequencing error gradients such as Illumina quality decay towards read ends and nanopore homopolymer errors would be concrete non-biological analogues for the same. The general condition is that any corruption that has a measurable position-specific or context-specific frequency profile qualifies, meaning the profile can be substituted for the mapDamage2 array. This is the design space DAM enables; the present work validates only the aDNA instance.

### 3.4 Fine-tuning Setup

For the base model DNABERT-2, a 117M-parameter BERT-style encoder was selected and pre-trained on multi-species genomic sequences using a tokenizer-free BPE architecture (Zhou et al., 2023). The chromosomal coordinates from gene_coords.csv were used to extract the CDS sequence for TRPV3, KCNK9, and HBB from GCF_024166365.1. After this, all sequences were converted to uppercase before the tokenisation, the reason being that the lowercase soft-masking notation from the RepeatMasker would cause DNABERT-2's BPE tokenisation to fall back to the character-level tokenisation, which would approximately double the token counts for repeat-rich loci such as KCNK9 (910 tokens mixed-case vs 400 tokens uppercase for the 2000 bp window). The sequences were then tiled into 2,000 bp sliding windows with a 200 bp stride and tokenised to a maximum of 512 tokens, which yielded 344 windows with TRPV3: 226, KCNK9: 97, HBB: 21 and was split 80/20 into training (275) and validation (69) sets (seed 42).

MLM was chosen as the baseline for the ablation design because it is the standard masking objective every existing DNA LM uses (Dalla-Torre et al., 2024; Ji et al., 2021; Nguyen et al., 2023; Zhou et al., 2023). Any alternative masking strategy, such as span, dynamic, or content-adaptive (Joshi et al., 2020; Levine et al., 2020; Liu et al., 2019) would change the granularity or schedule rather than spatial position, confounding the single variable under test, and zero-shot or random are reference floors (lower bounds), but MLM is the competitive incumbent. Thus, MLM is the only baseline that isolates spatial redistribution. DAM is a drop-in replacement for exactly that collator, matched at 15% C/G density, with the density being held fixed because the masking rate itself is a consequential design variable (Wettig et al., 2023). The design involved two fine-tuning runs performed under precisely identical conditions, utilising the same model, dataset, hyperparameters, and random seed, differing only in the data collator. The impact of the masking method, as the sole experimental variable, is isolated through this controlled ablation. Table 2 lists the hyperparameters, held identical across both runs.

Table 2. Hyperparameters shared identically across the DAM and MLM fine-tuning runs; the data collator's masking distribution is the sole differing variable. Spatial mask redistribution is thereby isolated as the experimental variable, with optimiser, schedule, batch, seed, and decay drift excluded by construction.

| **Parameter** | **Value** |
|---|---|
| Epochs | 20 |

| Batch size | 8 |
|---|---|
| Learning rate | 2 x $10^{-5}$ |
| LR schedule | Cosine with 10% warmup |
| Weight decay | 0.01 |
| Seed | 42 |

For the hardware, both runs were executed on Apple M1 Pro with MPS backend via HuggingFace Trainer (Wolf et al., 2020), where the wall-clock time for MLM was 11 hours and 24 minutes, whereas for DAM it was 14 hours and 32 minutes. All the training metrics, validation loss curves, and evaluation results were logged to Weights & Biases for full reproducibility (Biewald, 2020).

**3.5 In Silico Damage Simulation**

The C→T and G→A substitutions were applied in silico to the validation set windows utilising the empirical PMD frequencies from damage_profile.npy with the main objective of evaluating reconstruction performance under realistic degradation. For each nucleotide position $i$ in a validation window, a substitution event $X_i$ gets drawn independently from a Bernoulli distribution, which is then parameterised by the position-specific empirical damage frequency.

$$X_i \sim \text{Bernoulli}(p_i), \quad p_i = \begin{cases} \text{ct}_{5p}[\min(d_5(i),\, 69)] & \text{if nuc}(i) = \text{C} \\ \text{ga}_{3p}[\min(d_3(i),\, 69)] & \text{if nuc}(i) = \text{G} \\ 0 & \text{otherwise} \end{cases}$$

Equation [eq:bernoulli] represents the Bernoulli simulation equation, where $d_5(i)$ and $d_3(i)$ are the distances of the position $i$ from the 5’ and 3’ termini, respectively; nuc($i$) is the nucleotide at the position $i$, and $X_i = 1$ indicates a substitution event of either C→T or G→A. Additionally, positions carrying A or T receive zero substitution probability which is consistent with the deamination specificity of PMD, and all the Bernoulli draws use the seed 42 for reproducibility. Additionally, $p_i$ deliberately mirrors the structure of the masking probability, which was defined in Equation [eq:dam], ensuring that the evaluation directly probes the positions that DAM was trained to prioritise.

The BPE (Byte-Pair Encoding) tokenisation's segmentation is affected by neighboring characters(Schmidt et al., 2024), where substituting even a single nucleotide at position $i$ alters the token segment boundaries across a surrounding neighborhood, which produces a re-tokenised sequence that differs from the original positions far more than what was actually damaged. During the preliminary experiments, substituting 2 nucleotides would produce token-level differences at 31 positions, but upon re-tokenisation, masking all 31 positions would evaluate the reconstruction of approximately 150 nucleotides, whereas, in reality, only 2 were damaged, which introduced a systematic confound. To address this problem an algorithmic solution was used where, to isolate evaluation to exactly the damaged positions, the offset_mapping tensor from the original undamaged tokenisation was used for each damaged nucleotide i, the unique BPE token t(i) whose character span contains position i. Formally, the token index is determined by:

$$t(i) = k \quad \text{s.t.} \quad \text{start}_k \leq i < \text{end}_k$$

In the above equation, $start_k$ and $end_k$ are the character level offsets of the token k in the original sequence, which were further obtained from the tokenizer's offset_mapping output (Wolf et al., 2020). In the equation only tokens {t(i):$X_i$=1} were replaced with [MASK] in the original input_ids. Therefore, the damaged sequence was never re-tokenised, ensuring a bijective mapping between the damaged nucleotide positions and masked token positions.

From the simulation output, 43 out of the 69 validation windows contained at least one damaged nucleotide with a mean of $1.14 \pm 1.10$ damaged positions per window. Consistent across all the 43 affected windows, a total of 79 nucleotide positions were substituted. The sparse damage intensity is a reflection of the authentic empirical PMD rates, with a peak C→T frequency of 0.35% at position 1 and an exponential descent to background. Additionally, no synthetic inflation was implemented, thereby maintaining the ecological validity of the evaluation. Finally the resulting paired dataset of ground truth input_ids and masked input_ids was serialised to damaged_validation.npy with the intention of deterministic reuse across all subsequent evaluation scripts, ensuring that all the reported metrics are derived from identical damaged inputs.

### 3.6 Evaluation Protocol and Statistical Analysis

For the nucleotide recovery rate (NRR), the fraction of the masked positions at which the model's argmax prediction matches the ground truth token is:

$$\mathrm{NRR}(w) = \frac{|\{j \in \mathcal{M}_w : \hat{y}_j = y_j\}|}{|\mathcal{M}_w|}$$

As defined in Equation [eq:nrr], $\mathcal{M}_w$ is the set of masked positions in window $w$, with $\hat{y}_j = \mathrm{argmax}_v \mathrm{logit}_j(v)$ being the predicted token and $y_j$ signifying the ground truth token. In addition to this, there are two evaluation regimes: the background regime for C/G positions outside the terminal zone, testing whether DAM degrades specificity at positions it was not trained to prioritise; and the terminal regime for positions within the terminal zone, testing the reconstruction at damage-concentrated sites.

The terminal zone width in nucleotides is defined as $T_{END}$ where, for a window of length $L$, the terminal zone at the width $T_{END}$ comprises all the nucleotide positions which are within $T_{END}$ bases of either terminus and all C/G positions within this zone were deterministically masked. Additionally, at narrow terminal zones ($T_{END} = 3$), it was observed that the expected number of stochastic substitution events per window approaches zero given the empirical PMD rates, thus yielding a near-empty $\mathcal{M}_w$ for most of the windows and unreliable per-window statistics; deterministic masking guarantees a non-empty $\mathcal{M}_w$ at every $T_{END}$ value. We evaluate the system at discrete state values $T_{END} \in \{3,5,10,15,20,25\}$ alongside four baselines: zero-shot with the base as DNABERT-2 and no fine-tuning, frequency-matched random nucleotide prediction, fine-tuned MLM model, and fine-tuned DAM model. There are 626 windows across the 7 genes, of which 3 are training genes (TRPV3, KCNK9, and HBB) and 4 are held-out genes (TRPA1, UCP1, ADRB3, and FASN); and the held-out genes evaluate the model's generalisability beyond the fine-tuning distribution.

The evaluation setup is adequate to isolate the effect of DAM because the only free variable is the masking distribution, and everything else, such as model, data, seed, and hyperparameters, is identical. The 15% C/G density is matched as seen in Equation [eq:scale], and the background/terminal split is designed so that a localised effect can be distinguished from a general one. If DAM's advantage were just a general capability gain, it would have been seen in the background too; thus, the effect being attributable to spatial redistribution is mainly due to the two-regime evaluation. Together, these controls pin any DAM–MLM difference on spatial redistribution alone.

A two-sided paired $t$-test with windows as the pairing unit was performed, testing $H_0: \overline{NRR}_{DAM} = \overline{NRR}_{MLM}$, where the bootstrap confidence intervals for $\Delta(DAM - MLM)$ were computed from $B = 10{,}000$ resamples of windows with 95% CI alongside a significance threshold of $\alpha = 0.05$.

The ESMFold structural validation consists of DAM-reconstructed DNA translated to protein and submitted to the ESMFold via REST API (Lin et al., 2023) where the predicted structure aligns with the *Elephas maximus* reference via Kabsch superposition (Kabsch, 1976), scored by TM-score (Zhang & Skolnick, 2004) and C$\alpha$-RMSD across TRPV3, KCNK9, and HBB. For the biosecurity layer all the reconstructed windows passed through a 1D CNN classifier, which was trained on 10 virulence-associated gene classes, where the window gets flagged if the classifier's maximum class probability exceeds a fixed threshold; additionally, the AUC and flag rate are reported.

## 4. Results

### 4.1 Validation Loss Comparison

DAM achieved a validation loss of 3.274 against MLM's 3.757 at epoch 17, showing a 13% reduction where both the models converged and neither of them showed any overfitting, with the training and validation loss tracks remaining consistent, as seen in Figure 3.

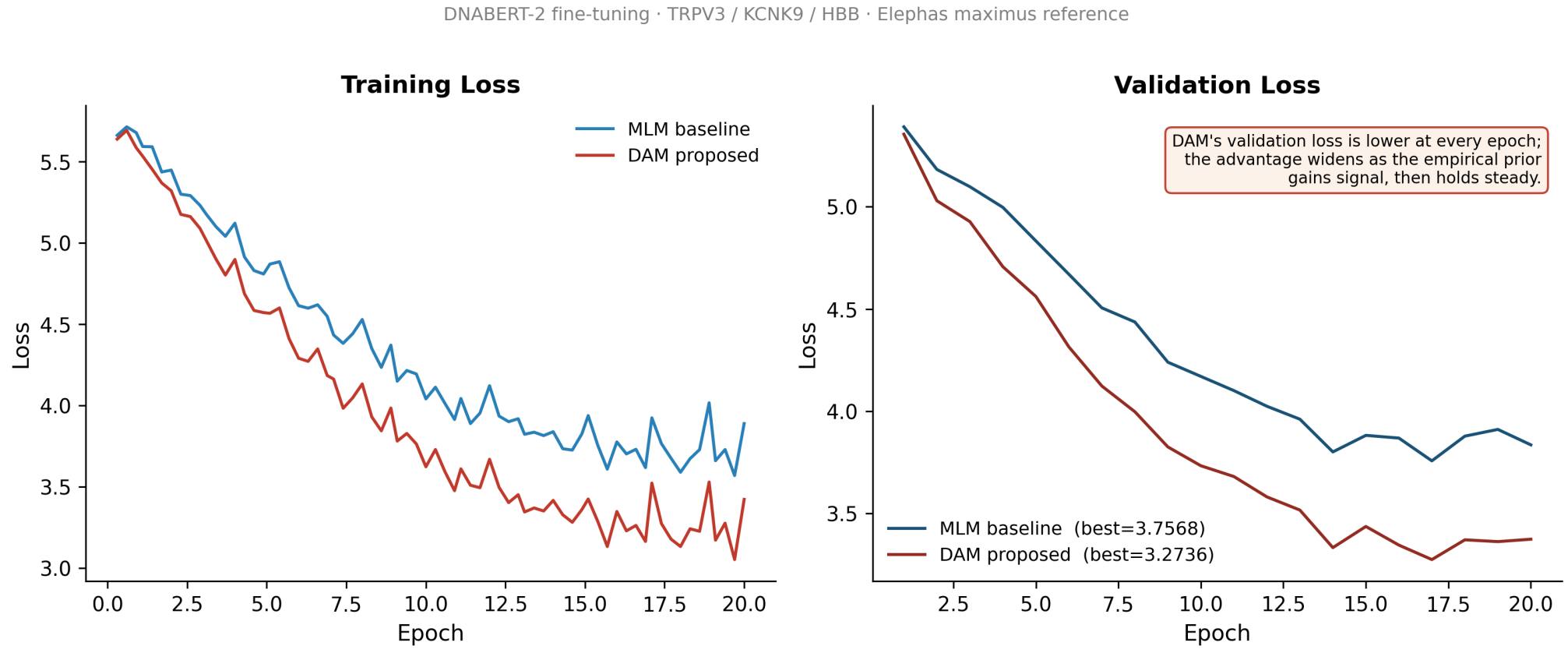

***Figure 3. DAM lowers validation cross-entropy throughout fine-tuning, evidencing that aligning the masking distribution with the empirical corruption profile yields a better-calibrated model - not merely a better argmax.*** *Training and validation loss curves for DAM and MLM over 20 epochs (best val: 3.274 vs. 3.757).*

There exists a loss reduction only in C/G positions where the masking density is matched between the DAM and MLM with both at 15% on C/G positions; thus, the gap is attributable only to the spatial redistribution of masking towards the terminal positions. This confirms that the concentrating masking positions present in the damage-prone termini force the model to learn a more specialised representation rather than distributing capacity uniformly across all positions.

### 4.2 Nucleotide Recovery: Background and Terminal Sites

Table 3 shows that at non-terminal C/G positions, both DAM and MLM are statistically indistinguishable, with NRR being 65.12% vs 65.70%, respectively, the mean difference being -0.58 pp, $t(42) = -0.14$, $p = 0.892$ and 95% CI $[-8.9, +7.8]$ pp, where the CI is symmetric about zero with no directional bias. DAM redistributes masking away from background positions, which, by design, should not affect background recovery; none was observed.

Table 3. Background nucleotide recovery rate across 43 damaged validation windows, evaluated at C/G positions outside the terminal zone. DAM and MLM do not differ significantly.

| **Metric** | **MLM Baseline** | **DAM Proposed** |
|---|---|---|
| NRR mean | 65.70% | 65.12% |
| NRR SD | 39.47% | 39.34% |
| Windows evaluated | 43 | 43 |
| *Paired statistics (DAM − MLM, $n = 43$)* | | |
| Mean difference | $-0.58$ pp | |
| $p$-value (two-sided) | 0.892 | |
| Bootstrap 95% CI | $[-8.9, +7.8]$ pp | |

At terminal positions ($T_{END} = 10$, all 69 validation windows, 325 C/G sites per model), DAM recovers 38.56% against MLM's 34.08% with the $\Delta = +4.48$ pp, $t(68) = 2.09$, $p = 0.041$ and 95% CI entirely positive $[+0.43, +8.69]$ pp as observed in Table 4, where CI entirely positive means that the lower bound excludes zero, which rules out a chance result in either direction. At the aggregate site level, DAM is correct at 124/325 positions against MLM's 109/325.

Table 4. Terminal-zone nucleotide recovery rate at $T_{END} = 10$ across all 69 validation windows (325 C/G sites). DAM significantly outperforms MLM; CI is entirely positive.

| Metric | MLM Baseline | DAM Proposed |
| --- | --- | --- |
| NRR mean | 34.08% | 38.56% |
| NRR SD | 27.70% | 27.24% |
| Correct / total sites | 109 / 325 | 124 / 325 |
| Windows evaluated | 69 | 69 |
| *Paired statistics* ($DAM - MLM$, $n = 69$) | | |
| Mean difference | $+4.48$ pp | |
| $t$-statistic | $t(68) = 2.09$ | |
| $p$-value (two-sided) | 0.041 | |
| Bootstrap 95% CI | $[+0.43, +8.69]$ pp | |

The null effect at the non-terminal positions shows a background/terminal dissociation, with a significant gain at the damage-concentrated zone, which confirms that DAM's recovery advantage is strictly localised to the region its masking gradient was designed to prioritise, which is consistent with Section 3.

### 4.3 Terminal Zone Sensitivity Analysis

At the position $T_{END} = 3$, it is observed that the four baselines rank ZeroShot (15.51%) $<$ MLM (20.49%) $<$ Random (27.67%) $<$ DAM (30.84%) alongside $\Delta(DAM - MLM) = +10.35$ pp with $p = 1.33 \times 10^{-10}$ and $n = 486$ windows with evaluable C/G positions. This clearly shows that MLM falls **below** random chance at $T_{END} = 3$ (20.49% against random chance's 27.67%). The standard fine-tuning with uniform masking actively underperforms a frequency-matched random predictor and thus fails to provide any useful inductive bias at peak PMD positions; DAM is the only method above random at this zone width as observed in Figure 4.

T_END = 3 nt · Innermost terminal zone
(Peak post-mortem damage — highest C→T probability)
MLM < Random: standard MLM produces a worse-than-random model at the very positions that matter most.
Random (theoretical 25%)
Nucleotide recovery rate
40%
35%
30%
25%
20%
15%
10%
5%
0%
MLM < random
15.5%
20.5%
27.7%
30.8%
+10.3 pp
(p < 0.001)
Zero-shot DNABERT-2
MLM fine-tuning
Random (chance)
DAM (proposed)
n = 486 windows · 852 terminal C/G sites · 7 genes (TRPV3, KCNK9, HBB, TRPA1, UCP1, ADRB3, FASN)

***Figure 4. Standard MLM produces a worse-than-random model at the very positions that matter most.*** *NRR at* $T_{END} = 3$*: MLM (20.49%) falls below random (27.67%); DAM (30.84%) is the only method above random.* $n = 486$ *windows,* $p = 1.33 \times 10^{-10}$.

Across all the six $T_{END} \in \{3,5,10,15,20,25\}$, DAM consistently outperforms all the three baselines at every width with all $p < 10^{-8}$ as observed in Table 5.

Terminal-zone NRR across $T_{END} \in \{3,5,10,15,20,25\}$, four baselines, 626 windows. DAM leads at every width; MLM falls below random at $T_{END} \leq 5$.

**Table 5.** Terminal-zone nucleotide recovery rate across $T_{END} \in \{3,5,10,15,20,25\}$, on 626 paired windows spanning seven genes. DAM outperforms MLM at every width (all < $10^{-8}$), with the gap peaking at $T_{END}$ ∈ (+10.35 pp) and narrowing as background positions dilute the terminal advantage.

| $T_{END}$ | Zero-Shot | MLM | Random | DAM | Δ (DAM−MLM) | *p* |
|---|---|---|---|---|---|---|
| 3 | 15.51% | 20.49% | 27.67% | 30.84% | +10.35 pp | $1.33 \times 10^{-10}$ |
| 5 | 19.24% | 24.02% | 25.69% | 31.18% | +7.17 pp | $9.09 \times 10^{-9}$ |
| 10 | 22.78% | 26.53% | 24.97% | 32.24% | +5.71 pp | $2.63 \times 10^{-10}$ |

| 15 | 24.46% | 27.98% | 25.77% | 32.16% | +4.18 pp | $9.36 \times 10^{-10}$ |
|---|---|---|---|---|---|---|
| 20 | 23.64% | 27.02% | 24.81% | 31.75% | +4.74 pp | $4.99 \times 10^{-14}$ |
| 25 | 23.74% | 26.73% | 25.65% | 31.64% | +4.91 pp | $1.41 \times 10^{-17}$ |

As shown in Figure 5, the DAM-MLM gap is largest at $T_{END} = 3$ (+10.35 pp) and decreases as the zone widens which is mechanistically consistent, where the wider zones tend to dilute the terminal-specific advantages with background C/G positions where both the models perform equivalently. Additionally, MLM exceeds random only from $T_{END} \geq 10$ onward and at $T_{END} \in \{3,5\}$ MLM remains below random, and this pattern holds across all the 7 genes, including the 4 held-out genes (TRPA1, UCP1, ADRB3, FASN) that were absent from fine-tuning, which confirms that the result is not specific to just the training gene distribution. Performance shows systematic improvement when the masking distribution follows the empirical degradation pattern, and when it does not, it falls towards or below the random baseline.

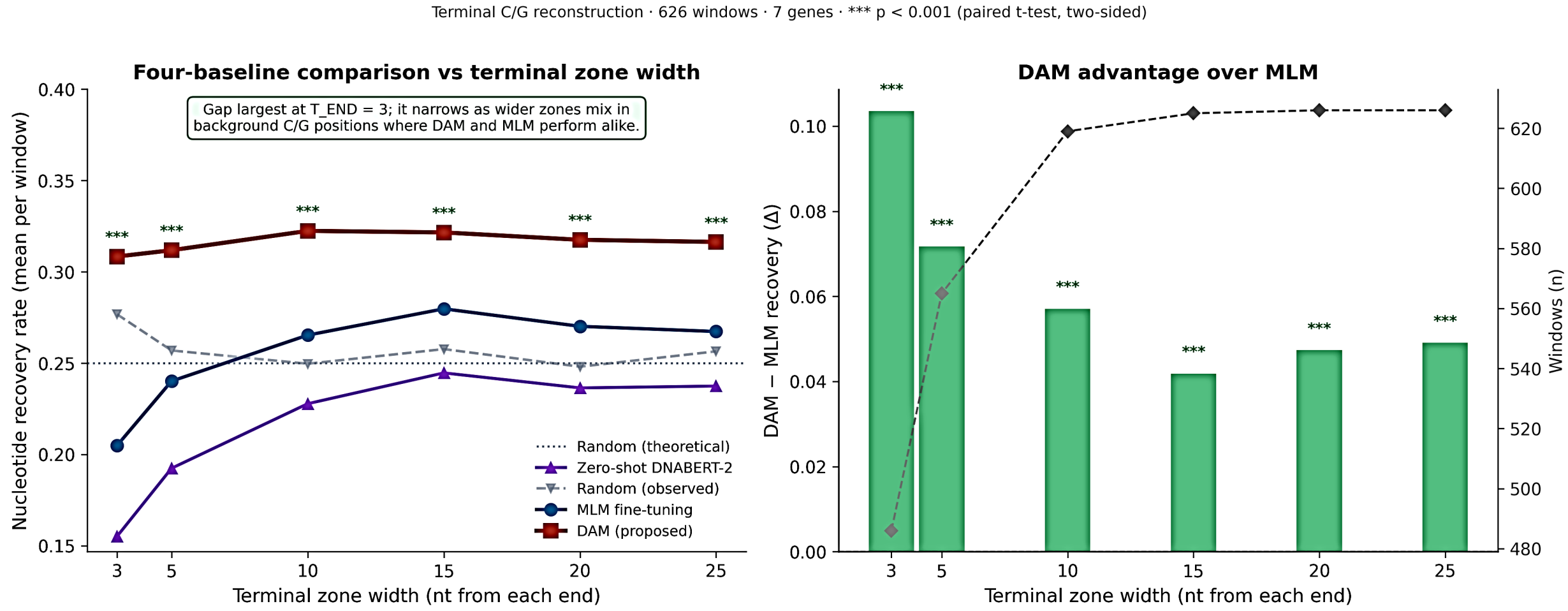

***Figure 5.*** *DAM's lead over MLM spans all six terminal-zone widths and is largest at innermost-terminal positions where damage peaks, confirming that directing training signal towards corrupted positions is beneficial even when those positions are a small minority of the fragment. Terminal-zone NRR vs.* $T_{END}$ *for four baselines across 626 windows. Δ(−) peaks at* $T_{END} = 3$ *(+10.35 pp) and narrows as wider zones dilute the terminal advantage.*

Figure 6 shows the DAM advantage is sharpest at $d = 1$ where DAM recovers 31.25% against MLM's 6.25% at the 5' terminus, with the gap decaying monotonically towards the interior positions.

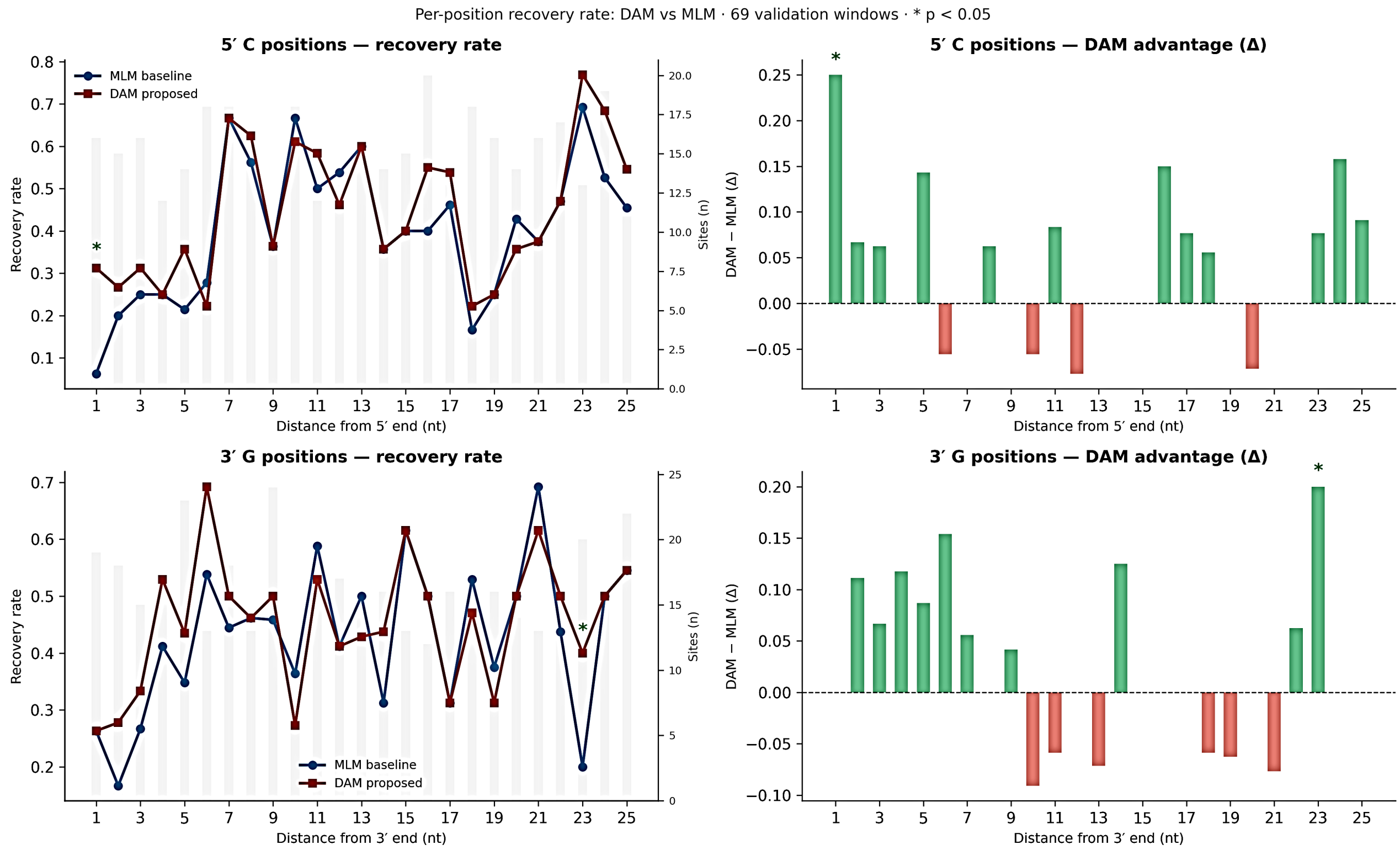

***Figure 6**. Per-position NRR for DAM and MLM at read-end distances $d = 1$–$10$ (5'-C and 3'-G). DAM advantage is sharpest at $d = 1$: 31.25% vs. 6.25%.*

The damage intensity sweep (Figure 7) confirms the advantage scales with PMD severity at simulated peak rates $\geq 20\%$, $\Delta > 11$ pp and $p < 0.01$ and at 5% the effect is not significant and is directionally consistent ($p = 0.161$, $n = 31$) which was expected given the low site count.

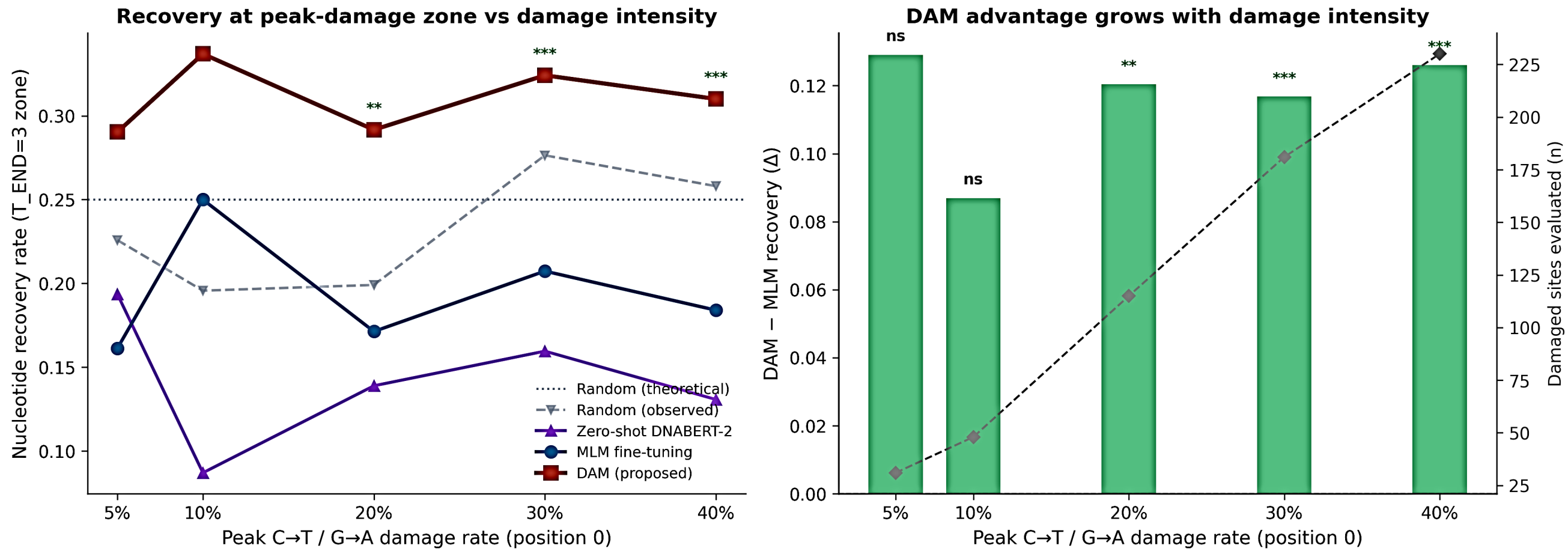

***Figure** 7. NRR vs. simulated peak PMD rate (5%–40%). DAM advantage is significant at peak rates* $\geq 20\%$ *(*$p < 0.01$*); directionally consistent but not significant at 5% (*$p = 0.161$*,* $n = 31$*).*

## 4.4 Structural Validation via ESMFold

A structural validation check is not a separate result but a check that ensures the recovery rate improvement truly produces valid proteins and that the fold is not damaged by either masking method. To confirm that the nucleotide level reconstruction errors are not propagated to the fold-level disruptions, the CDS sequences were reconstructed from both models across the three training genes (TRPV3, KCNK9, and HBB) and were submitted to ESMFold (Lin et al., 2023) via the public REST API for ab initio protein structure prediction. The folding was restricted to the N-terminal domain with residues 1-400, mainly because the full-length TRPV3 protein (791 aa) had exceeded the ESMFold API residue limit. The damage was applied at 10-30 times the authentic PMD rates for this phase, which amplified it beyond the empirical frequencies used in Sections 4.1 to 4.3, to stress test the structural robustness under conditions which far exceeded those of the actual mammoth specimens, and thus the fold integrity under this application is sufficient to also guarantee integrity under the authentic degradation.

Two structurally-similar metrics, TM-score (Zhang & Skolnick, 2004) and C$\alpha$-RMSD after Kabsch superposition (Kabsch, 1976) were computed between each reconstructed structure and *Elephas maximus* reference fold.

$$\text{TM-score} = \max\left[\frac{1}{L_{\text{ref}}}\sum_{i=1}^{L_{\text{ali}}}\frac{1}{1+(d_i/d_0)^2}\right], \quad d_0(L_{\text{ref}}) = 1.24\sqrt[3]{L_{\text{ref}}-15} - 1.8$$

In Equation [eq:tmscore] $d_i$ is the C$\alpha$ distance of the $i$-th aligned residue pair after superposition, $L_{\text{ref}}$ is the reference length, and $L_{ali}$ is the aligned residues. Additionally, the TM-score is normalised by $L_{\text{ref}}$, where scores above 0.5 indicate the same fold topology and above 0.9 indicate near-identical structures.

$$\text{RMSD} = \sqrt{\frac{1}{n}\sum_{i=1}^{n}\left\|r_i - r_i'\right\|^2}$$

The C$\alpha$ coordinates of the $i$-th aligned residue pair for Equation [eq:rmsd] are $r_i$ and $r_i'$ following the application of the optimal rotation matrix from Kabsch superposition.

From Table 6, the per-gene structural metrics can be understood, where for TRPV3 with 400 residues both models yield TM-score = 0.9808, C$\alpha$-RMSD = 1.092 Å, $\text{pLDDT}_{\text{recon}}$ = 77.44 vs. reference 77.37, making it structurally identical across both MLM and DAM; for KCNK9 with 347 residues both models yield TM-score = 0.9520, C$\alpha$-RMSD = 1.732 Å, $\text{pLDDT}_{\text{recon}}$ = 67.02 vs. reference 67.94 making it also identical; and for HBB with 162 residues MLM gives a TM-score = 0.9728, RMSD = 0.815 Å, $\text{pLDDT}_{\text{recon}}$ = 80.78, and DAM gives TM-score = 0.9714, RMSD = 0.840 Å, $\text{pLDDT}_{\text{recon}}$ = 81.34, leading to a difference of ΔTM = 0.0014 and ΔRMSD = 0.025 Å, which is well within the ESMFold stochasticity. All of these are shown in Figure 8, where the per-residue pLDDT profiles confirm this agreement across all three genes.

ESMFold structural validation. C$\alpha$-RMSD after Kabsch superposition; TM-score normalised by reference length. TRPV3 folded on N-terminal domain (residues 1–400).

Table 6. ESMFold structural validation matrix across the three training genes (TRPV3, KCNK9, HBB) at PMD rates amplified 10–30× beyond authentic frequencies, evaluated by TM-score (after Kabsch superposition), Cα-RMSD, and pLDDT (reference vs. reconstruction). All six reconstructions exceed TM-score > 0.95 — well above the same-fold (0.5) and near-identical (0.9) thresholds - with TRPV3 and KCNK9 identical between masking strategies and HBB's ΔTM = 0.0014, well within ESMFold stochasticity.

| Gene | Method | $pLDDT_{ref}$ | $pLDDT_{recon}$ | TM-score | RMSD (Å) |
|---|---|---|---|---|---|
| TRPV3 | MLM | 77.37 | 77.44 | 0.9808 | 1.092 |
| TRPV3 | DAM | 77.37 | 77.44 | 0.9808 | 1.092 |
| KCNK9 | MLM | 67.94 | 67.02 | 0.9520 | 1.732 |
| KCNK9 | DAM | 67.94 | 67.02 | 0.9520 | 1.732 |
| HBB | MLM | 79.70 | 80.78 | 0.9728 | 0.815 |
| HBB | DAM | 79.70 | 81.34 | 0.9714 | 0.840 |

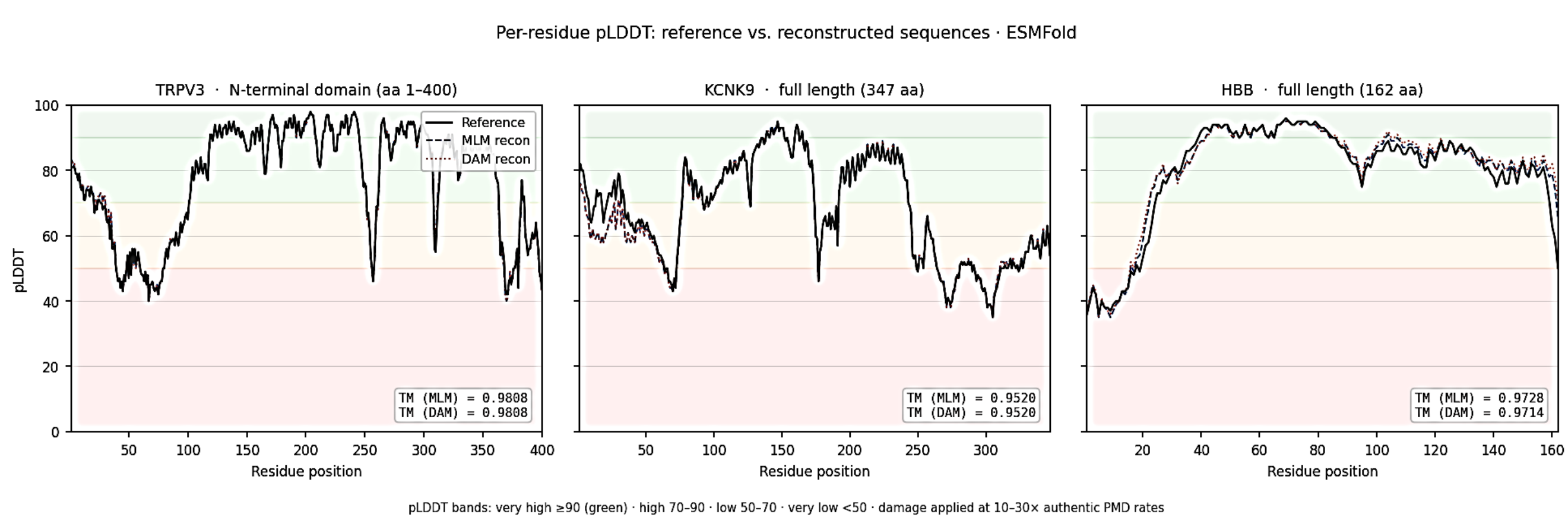

***Figure 8. ESMFold reconstructions achieve TM-score*** $>0.95$ ***across all six reconstructions (three genes) even under synthetic damage amplified 10–30× beyond authentic PMD rates, evidencing that the sequence-level reconstruction gain translates into biologically plausible structures.*** *Per-residue pLDDT for reference and reconstructed sequences across TRPV3, KCNK9, and HBB. MLM and DAM reconstructions track the reference profile in all three genes; TM-scores are annotated per panel.*

Therefore, all six reconstructions achieve a TM-score $>0.95$, which is well above the same-fold (0.5) and near-identical (0.9) thresholds, and thus the fold topology is fully preserved under amplified damage in every single case. Given that MLM's and DAM's outputs for TRPV3 and KCNK9 are structurally equivalent, this indicates that the nucleotide-level differences between the models fall exclusively at synonymous or non-coding positions, thus producing no change in the translated amino acid sequence, and the HBB divergence ($\Delta$TM = 0.0014) is negligible. These results confirm that even under degradation far exceeding the authentic PMD frequencies, the recovery gains do not damage the protein fold, and both models produce structurally valid output.

**4.5 Biosecurity Classification**

Screening is a necessary part of the pipeline, not an optional one, because any system that produces CDS sequences for downstream usage must have a dual-use risk check (Diggans & LeProust, 2019; Peccoud et al., 2017) The layer is made up of a 1D CNN trained on 300 bp sliding windows where the positive training set consists of 20 pathogen virulence sequences across 11 gene classes, namely *E. coli stx1/stx2, S. aureus mecA/spa, Listeria hlyA, Salmonella sipA/invA, V. cholerae ctxA/ctxB,* and *P. aeruginosa exoS/exoU* (n = 266 positive windows, 626 negative). A window is flagged if the classifier's maximum class probability exceeds a fixed decision threshold. Two primary metrics are reported: AUC on the held-out validation set and flag rate on all 626 reconstructed windows.

From the results, the performance in the validation set, AUC = 0.935, accuracy = 86.6%, precision = 78.4%, and recall = 75.5%, confirms that the classifier is discriminative before applying it to the reconstructed windows. Of the 626 reconstructed windows, only 11 were flagged (flag rate = 1.76%) and 98.2% were cleared. The gene-level flags were 6 for TRPA1, 3 for FASN, 1 for ADRB3, and 1 for HBB; all 11 flags are present in the Asian elephant reference genome before any reconstruction and can be observed in Figure 9.

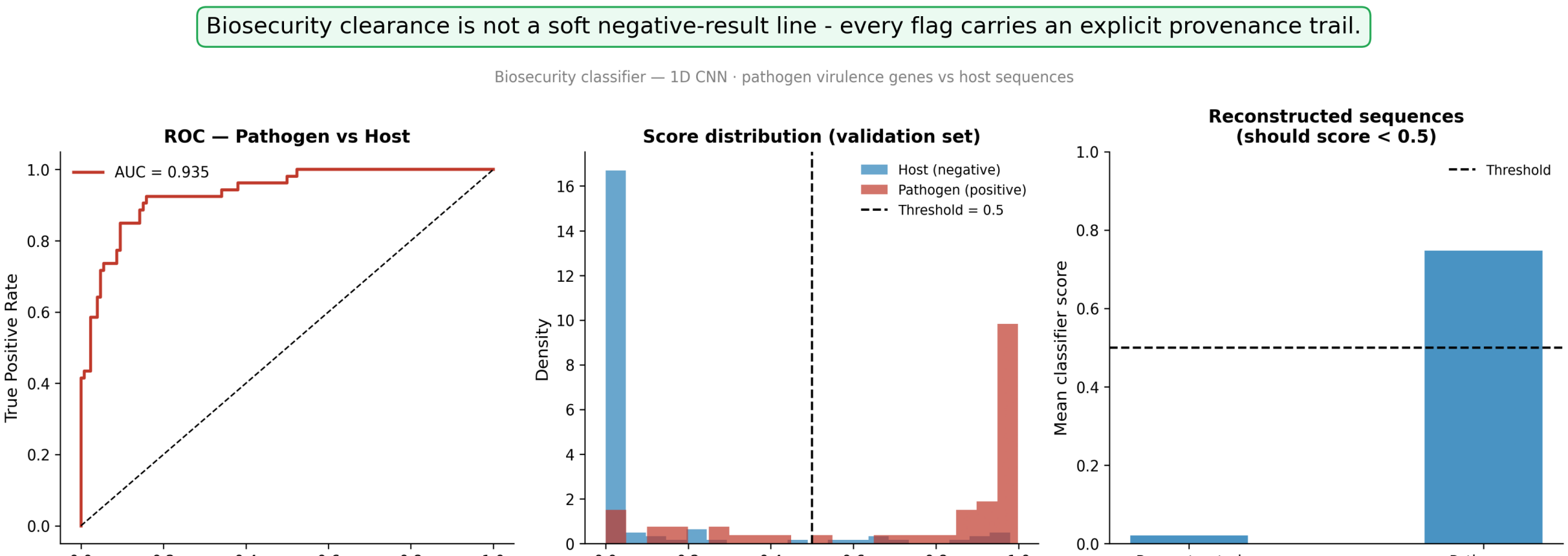

***Figure 9. The integrated biosecurity layer clears 98.2% of reconstructed windows with*** $AUC = 0.935$***, and every flag traces back to reference-genome features rather than to reconstruction artefacts.*** *Biosecurity classifier output across 626 reconstructed windows.* $AUC = 0.935$ *on held-out validation; 11/626 windows flagged (1.76%), all attributable to reference-genome features rather than reconstruction artefacts.*

In the TRPA1 flags, ankyrin repeat domains are structurally conserved between bacteria and eukaryotes, making the classifier react to the repeat motif and not a pathogenic sequence. Additionally, the high-GC windows (GC = 0.637) for FASN flags are compositionally similar to bacterial CDS, which leads to false-positive results based on base composition rather than sequence homology. Ultimately, neither MLM nor DAM reconstruction introduced any kind of novel virulence-gene signatures, and the reconstruction does not elevate the biosecurity risk above the baseline reference genome. The 1.76% flag rate establishes an upper-bound screen, and VESTIGE's outputs are safe for downstream paleogenomics pipelines, completing the end-to-end system by guaranteeing that the outputs are safe to release, unlike reconstruction approaches that report only accuracy.

## 5. Discussion

It is observed that at $T_{END} = 3$, MLM recovers 20.49% of terminal nucleotides, which is below the 27.67% achievable by a frequency-matched random predictor, and this is because the uniform masking during training gives the model no incentive to specialise at the 5' and 3' termini where the deamination is concentrated, and thus the model instead learns a representation shaped by interior positions and is actively wrong at peak PMD sites. DAM at the same zone width recovers 30.84%, showing that redirecting the masking gradient to empirically damage-concentrated positions produces a meaningful and statistically robust improvement ($p = 1.33 \times 10^{-10}, n = 486$). As observed in Figure 5, $\Delta(DAM - MLM)$ peaks at $T_{END} = 3$ (+10.35 pp) and narrows monotonically as the terminal zone widens which is mechanistically expected because the wider zone also includes interior C/G positions where both the models perform equivalently with the background $p = 0.892$ as observed in Table 3, diluting the terminal-specific advantage.

There is a 13% cross-entropy reduction (3.274 vs 3.757), which is consistent with the NRR results but measures a qualitatively different quantity. Furthermore, the CE loss penalises the full predictive distribution over the vocabulary, while the NRR captures only the argmax, a lower-CE model that assigns a higher probability mass to the correct token without necessarily changing the argmax, particularly when the correct nucleotide competes with a near-identical BPE token; also, this distinction is well-established in the MLM masking literature (Wettig et al., 2023). Therefore, the loss improvement shows a sharper and better-calibrated distribution at the masked C/G positions, which is also a prerequisite for the downstream uses that consume token probabilities directly, such as variant scoring or ancestral sequence inference, and the binary correct/incorrect metric of NRR alone fails to fully capture it.

The nucleotide-level reconstruction differences between DAM and MLM do not propagate to fold-level disruption, which is confirmed by the ESMFold structural validation, where all six reconstructed sequences achieve TM-score $> 0.95$ vs the *Elephas maximus* reference and are above both the same-fold (0.5) and near-identical-structure (0.9) thresholds. For TRPV3 and KCNK9, the outputs produced are structurally indistinguishable between the two models, and the nucleotide-level differences fall exclusively at synonymous or non-coding positions. The HBB divergence ($\Delta$TM = 0.0014) falls within ESMFold stochasticity; the structural validation performed was 10-30 times the authentic PMD rates, where the fold integrity preserved under amplified damage guarantees integrity even under the authentic mammoth degradation. The biosecurity classifier independently confirms that reconstruction does not elevate dual-use risk above the baseline reference genome, where all the 11 flagged windows are attributable to reference-genome features (conserved ankyrin repeats in TRPA1, high-GC composition in FASN) and not reconstruction artefacts. The 98.2% clearance rate shows that VESTIGE's outputs are safe for downstream paleogenomics pipelines.

## 6. Generalisability

### 6.1 Post-Mortem Damage Beyond Mammoth

The core finding - that aligning the masking distribution with the empirical corruption profile produces a meaningful and statistically robust improvement at corrupted positions - is not specific to mammoth as a species or to cytosine deamination as a chemistry. Other ancient and historical specimens carry characterised damage signatures (Orlando et al., 2021),(Shapiro & Hofreiter, 2014),(Dabney et al., 2013) that map directly onto the DAM mechanism. The mammoth parallel extends most naturally to cave bear, aurochs, archaic *Homo*, and permafrost-preserved plant material, where the fragment-length and deamination regime are quantitatively similar and the profile can be supplied by the same mapDamage2 pipeline. The empirical decay constants vary across preservation conditions, but the *form* of the terminal gradient is conserved, and substituting a new profile requires no code change - a property inherited from the abstract-collator interface established in Section 3.

### 6.2 Biological Damage Beyond PMD

DNA extracted from formalin-fixed paraffin-embedded tumour biopsies (Do & Dobrovic, 2014) and from herbarium/museum specimens(Porrelli et al., 2026) carries its own C→T-dominant degradation signal with a longer-tailed decay and is routinely discarded or post-hoc corrected by trimming. The same PMD array interface accepts an FFPE-tuned frequency profile directly: substituting the FFPE-specific profile for the aDNA PMD array yields a DAM instance that prioritises reconstruction at the exact positions where the formalin-derived lesions concentrate. Bisulfite-conversion data - the inverse problem, in which every unmodified C is converted to U/T - is digitally equivalent to a prior that places a *high* masking mass on every C position that is *not* in a CpG island, and would likewise feed straight into the DAM density-matched weighting without code modification.

### 6.3 Quality-Conditioned and Platform-Derived Priors

For noisy genomic data of the kind generated in low-coverage, environmental, or shotgun metagenomic sequencing (Quince et al., 2017), the corruption is not strictly position-deterministic but follows characterisable quality-scored and context-dependent distributions. The same collator accepts a quality-conditioned masking prior in place of a position-conditioned prior, with no algorithmic modification required: each token's probability of being selected is drawn from its Phred-scaled likelihood of containing a base-calling error. Platform-specific error profiles - Illumina quality decay toward read ends(Foox et al., 2021), nanopore homopolymer-deletion frequencies (Rang et al., 2018), and PacBio indel concentrations(Foox et al., 2021) - are routinely quantified during QC and become plug-and-play masking priors. The collator treats any vector with the same shape as $\mathrm{ct}_{5p}$ / $\mathrm{ga}_{3p}$ as valid, which makes the approach container-type-agnostic rather than biological-specific.

### 6.4 Reframing DAM as an Intelligent-Systems Training Routine

VESTIGE is also a candidate routine for expert and knowledge-based systems whose inputs are corrupted according to a measurable profile. Examples include: clinical decision support systems operating on OCR-degraded patient records; legal expert systems on corrupted historical archives; industrial fault detection systems on noisy sensor streams where error profiles exhibit directional bias. The DAM collator generalises to any sequence-input expert system whose empirical corruption distribution is known per-position.

Taken together, the analogues share a single operational principle: **align the self-supervised masking objective with the empirical corruption distribution of the deployment setting**. This positioning is consistent with the masking-strategy lineage reviewed in section 2

(Devlin et al., 2019; Joshi et al., 2020; Levine et al., 2020; Liu et al., 2019; Tang et al., 2026) and inherits one essential property from the DAM design - the pre-trained model weights are 100% retained; only the supervised signal is rerouted. Unlike domain-adaptation approaches that require continued or extended pre-training, DAM operates as a collator-level intervention that does not touch the encoder. This makes DAM a candidate knowledge-guided training routine for intelligent systems operating on degraded or noisy sequence inputs in a manner broader than genomics alone: any empirical per-position, per-quality, or per-context corruption profile that has measurable structure qualifies as a DAM instance. Cross-domain quantitative validation across FFPE, metagenomic, and nanopore-rescued benchmarks is the next experimental step.

## 7. Limitations and Future Work

The DAM collator is calibrated to only one set of the PMD profiles against one reference genome; thus, there is a reference bias introduced due to alignment to a single reference, which is inherent to short-read mapping (Günther & Nettelblad, 2019); the generalisability to other aDNA systems, like taxonomically distant species pairs, still remains to be established. Additionally, the training set comprises 344 windows across three genes, where the $T_{END}$ sweep evaluates 626 windows across seven genes, including four held-out genes, but the total sequence diversity is still narrow by modern LM standards. The empirical validation is confined to a single domain (mammoth aDNA under post-mortem damage), and thus the cross-domain extensions discussed above (FFPE, metagenomic, and sequencing-error data) remain conceptual and untested. Due to the ESMFold REST API residue limit, the structural validation was restricted to residues 1-400, where the full-length validation for TRPV3 (791 aa) was not possible under the current setup.

One of the most direct extensions to this would be curriculum masking, where one would begin fine-tuning with standard MLM to build broad sequence representations and then progressively increase the damage-gradient weight of DAM as the training proceeds. Previously, these types of curriculum strategies have shown consistent benefits in sequence modelling tasks (Bengio et al., 2009) and could help with the cold-start disadvantage that DAM faces before the model has any prior knowledge about terminal positions. A multi-reference alignment could be integrated, whereby applying VESTIGE in multiple closely related reference genomes would reduce reference bias and allow damage-aware masking across a richer sequence context. Also, extending this pipeline to other well-preserved aDNA datasets like cave bear, aurochs, and ancient human would help test whether the MLM-below-random failure mode is universal at the peak PMD sites or is just specific to the mammoth/elephant divergence time and the damage regime studied here.

**8. Conclusion**

Standard masked-language-model fine-tuning applies a uniform 15% masking probability across every token position, a position-agnostic assumption that fails the moment the corruption process is known, characterised, and concentrated at predictable positions. The empirical finding of VESTIGE is captured in a single sentence: at peak post-mortem-damage sites (the innermost terminal zone, $T_{\mathrm{END}} = 3$, with peak C→T rates at $d = 1$), standard MLM fine-tuning of DNABERT-2 recovers only 20.49% of damaged nucleotides - below the 27.67% achievable by a frequency-matched random predictor - whereas VESTIGE achieves 30.84% ($p = 1.33 \times 10^{-10}$, $n = 486$). Position-agnostic masking is not a neutral choice; it is actively harmful when the corruption is concentrated and known, and damage-informed masking is therefore a prerequisite for useful reconstruction, not an optional refinement.

The result generalises across six terminal-zone widths $T_{\mathrm{END}} \in \{3,5,10,15,20,25\}$: DAM leads MLM at every width, with the per-window mean difference $\Delta = +4.18$ to $+10.35$ percentage points (all $p < 10^{-8}$, $n = 626$ paired evaluation windows spanning two specimens and seven genes, including four held out from fine-tuning). Validation cross-entropy falls by 13% (3.274 vs. 3.757), ESMFold reconstructions superposed against the *Elephas maximus* reference via Kabsch rotation achieve TM-score > 0.95 across all six reconstructions (three genes) even under PMD simulation amplified 10×–30× beyond the empirical profile, and the integrated 1D CNN biosecurity classifier clears 98.2% of reconstructed windows (11 of 626 flagged) with AUC = 0.935 on a held-out virulence-gene test set, every flag attributable to reference-genome features (ankyrin repeats in TRPA1, high-GC composition in FASN) rather than to reconstruction artefacts. By combining three validation layers - reconstruction accuracy, structural plausibility, and biosecurity clearance - in a single pipeline, VESTIGE distinguishes itself from prior DNA LM work that reports only one of the three.

The design principle is domain-agnostic. An empirical corruption profile with measurable position-, quality-, or context-specific structure qualifies as a VESTIGE instance; only the prior vector changes. Concrete analogues already map onto the existing collator interface: FFPE tumour-specimen damage supplies a longer-tailed C→T frequency array; bisulfite-conversion data supplies a context-specific CpG-masked prior; low-coverage and shotgun metagenomic noise supplies a quality-conditioned prior; and nanopore homopolymer-error profiles supply platform-derived priors. VESTIGE is therefore most naturally read not as a paleogenomics tool but as a knowledge-guided training routine for intelligent systems operating on degraded or noisy sequence inputs, alongside the masking-strategy lineage that spans dynamic, span, pointwise-mutual-information, and multi-strategy schedules.

The aDNA instantiation reported here is fully validated; the projected analogues are not yet empirically validated and constitute the natural next experimental step. Curriculum variants - in which the DAM weight increases progressively from a standard-MLM warm-up - and multi-reference alignment variants are the most direct extensions. By releasing the complete VESTIGE pipeline as an open-source artefact, we aim to make corruption-aware fine-tuning a deployable, drop-in component of any genomic LM training stack that operates on degraded or noisy sequence inputs.

**Funding**

This research did not receive any specific grant from funding agencies in the public, commercial, or not-for-profit sectors.

**Declaration of Competing Interest**

The authors declare no competing interests.

**Author contributions: CRediT (left blank for review)**

**Declaration of generative AI and AI-assisted technologies in the manuscript preparation process**

During the preparation of this work the authors used generative AI in order to assist with manuscript structuring and editorial feedback. After using this tool/service, the author reviewed and edited the content as needed and takes full responsibility for the contents of the published article.

**Data Availability**

There are 3 primary data sources: (1) Raw aDNA reads, which are publicly available on NCBI SRA under accessions ERR855944 and ERR852028 (Palkopoulou et al. 2015); reference genome from NCBI RefSeq accession GCF_024166365.1 (EanMak 1.0); (2) All the code, checkpoints and evaluation data are available at https://github.com/Ganglet/Vestige, Zenodo DOI archived at https://doi.org/10.5281/zenodo.20601069; (3) The training logs and metrics are available at W&B project.

**References**

Bengio, Y., Louradour, J., Collobert, R., & Weston, J. (2009). *Curriculum learning*. https://doi.org/10.1145/1553374.1553380

Biewald, L. (2020). *Experiment Tracking with Weights and Biases*. https://www.wandb.com/

Briggs, A. W., Stenzel, U., Johnson, P. L., Green, R. E., Kelso, J., Prüfer, K., Meyer, M., Krause, J., Ronan, M. T., Lachmann, M., & Pääbo, S. (2007). Patterns of damage in genomic DNA sequences from a Neandertal. *Proceedings of the National Academy of Sciences*, *104*(37), 14616–14621. https://doi.org/10.1073/pnas.0704665104

Dabney, J., Meyer, M., & Paabo, S. (2013). Ancient DNA Damage. *Cold Spring Harbor Perspectives in Biology*, *5*(7). https://doi.org/10.1101/cshperspect.a012567

Dalla-Torre, H., Gonzalez, L., Mendoza-Revilla, J., Carranza, N. L., Grzywaczewski, A., Oteri, F., Dallago, C., Trop, E., Almeida, B. P. de, Sirelkhatim, H., Richard, G., Skwark, M. J., Beguir, K., Lopez, M., & Pierrot, T. (2024). *Nucleotide Transformer: building and evaluating robust foundation models for human genomics*. https://doi.org/10.1038/s41592-024-02523-z

Devlin, J., Chang, M., Lee, K., & Toutanova, K. (2019). https://doi.org/10.18653/v1/n19-1423

Diggans, J., & LeProust, E. M. (2019). *Next Steps for Access to Safe, Secure DNA Synthesis*. https://doi.org/10.3389/fbioe.2019.00086

Do, H., & Dobrovic, A. (2014). *Sequence Artifacts in DNA from Formalin-Fixed Tissues: Causes and Strategies for Minimization*. https://doi.org/10.1373/clinchem.2014.223040

Foox, J., Tighe, S., Nicolet, C. M., Zook, J. M., Byrska-Bishop, M., Clarke, W. E., Khayat, M. M., Mahmoud, M., Laaguiby, P. K., Herbert, Z. T., Warner, D., Grills, G. S., Jen, J., Levy, S., Xiang, J., Alonso, A., Zhao, X., Zhang, W., Teng, F., … Mason, C. E. (2021). Performance assessment of DNA sequencing platforms in the ABRF Next-Generation Sequencing Study. *Nature Biotechnology*, *39*(9), 1129–1140. https://doi.org/10.1038/s41587-021-01049-5

Ginolhac, A., Rasmussen, M., Gilbert, M. T. P., Willerslev, E., & Orlando, L. (2011). *mapDamage: testing for damage patterns in ancient DNA sequences*. https://doi.org/10.1093/bioinformatics/btr347

Günther, T., & Nettelblad, C. (2019). *The presence and impact of reference bias on population genomic studies of prehistoric human populations*. https://doi.org/10.1371/journal.pgen.1008302

Ji, Y., Zhou, Z., Liu, H., & Davuluri, R. V. (2021). *DNABERT: pre-trained Bidirectional Encoder Representations from Transformers model for DNA-language in genome*. https://doi.org/10.1093/bioinformatics/btab083

Jónsson, H., Ginolhac, A., Schubert, M., Johnson, P. L., & Orlando, L. (2013). *mapDamage2.0: fast approximate Bayesian estimates of ancient DNA damage parameters*. https://doi.org/10.1093/bioinformatics/btt193

Joshi, M., Chen, D., Liu, Y., Weld, D. S., Zettlemoyer, L., & Levy, O. (2020). *SpanBERT: Improving Pre-training by Representing and Predicting Spans*. https://doi.org/10.1162/tacl_a_00300

Jumper, J., Evans, R., Pritzel, A., Green, T., Figurnov, M., Ronneberger, O., Tunyasuvunakool, K., Bates, R., Žídek, A., Potapenko, A., Bridgland, A., Meyer, C., Köhl, S., Ballard, A. J., Cowie, A., Romera-Paredes, B., Nikolov, S., Jain, R., Adler, J., … Hassabis, D. (2021). Highly accurate protein structure prediction with AlphaFold. *Nature*, *596*(7873), 583–589. https://doi.org/10.1038/s41586-021-03819-2

Kabsch, W. (1976). *A solution for the best rotation to relate two sets of vectors*. https://doi.org/10.1107/s0567739476001873

Lefeuvre, M. (2023). *MaelLefeuvre/pmd-mask: pmd-mask v0.3.2*. https://doi.org/10.5281/zenodo.15403155

Levine, Y., Lenz, B., Lieber, O., Abend, O., Leyton-Brown, K., Tennenholtz, M., & Shoham, Y. (2020). *PMI-Masking: Principled masking of correlated spans*. https://doi.org/10.48550/arxiv.2010.01825

Li, H. (2013). *Seqtk: Toolkit for Processing Sequences in FASTA/q Formats*. https://github.com/lh3/seqtk

Li, H., & Durbin, R. (2009). *Fast and accurate short read alignment with Burrows–Wheeler transform*. https://doi.org/10.1093/bioinformatics/btp324

Lin, Z., Akin, H., Rao, R., Hie, B., Zhu, Z., Lu, W., Smetanin, N., Verkuil, R., Kabeli, O., Shmueli, Y., Costa, A. dos S., Fazel-Zarandi, M., Sercu, T., Candido, S., & Rives, A. (2023). *Evolutionary-scale prediction of atomic-level protein structure with a language model*. https://doi.org/10.1126/science.ade2574

Liu, Y., Ott, M., & Goyal, N. (2019). *RoBERTa: A Robustly Optimized BERT Pretraining Approach*. https://arxiv.org/abs/1907.11692

Lynch, V. J., Bedoya-Reina, O. C., Ratan, A., Sulak, M., Drautz-Moses, D. I., Perry, G. H., Miller, W., & Schuster, S. C. (2015). *Elephantid Genomes Reveal the Molecular Bases of Woolly Mammoth Adaptations to the Arctic*. https://doi.org/10.1016/j.celrep.2015.06.027

Nguyen, É., Poli, M., Faizi, M., Thomas, A. W., Birch-sykes, C. J., Wornow, M., Patel, A., Rabideau, C. M., Massaroli, S., Bengio, Y., Ermon, S., Baccus, S. A., & Ré, C. (2023). *HyenaDNA: Long-Range Genomic Sequence Modeling at Single Nucleotide Resolution*. https://doi.org/10.48550/arxiv.2306.15794

Orlando, L., Allaby, R. G., Skoglund, P., Sarkissian, C. D., Stockhammer, P. W., Ávila-Arcos, M. C., Fu, Q., Krause, J., Willerslev, E., Stone, A. C., & Warinner, C. (2021). Ancient DNA analysis. *University of Southern Denmark Research Portal (University of Southern Denmark)*, *1*(1). https://doi.org/10.1038/s43586-020-00011-0

Palkopoulou, E., Mallick, S., Skoglund, P., Enk, J., Rohland, N., Li, H., Omrak, A., Vartanyan, S., Poinar, H. N., Götherström, A., Reich, D., & Dalén, L. (2015). *Complete Genomes Reveal Signatures of Demographic and Genetic Declines in the Woolly Mammoth*. https://doi.org/10.1016/j.cub.2015.04.007

Panis, D. D., Arantes, L. S., Brown, T., Somenzi, E., Wibbelt, G., Ballaco, J., Mountcastle, J., Brajuka, N., Joardar, V., Fédrigo, O., Thibaud-Nissen, F., Ryder, O. A., Jarvis, E., Lummaa, V., & Mazzoni, C. J. (2026). *The reference genome of the Asian Elephant (Elephas maximus): a foundation for conservation and genomic research*. https://doi.org/10.1186/s12864-026-12821-9

Park, J., Park, H., Jeong, E., & Teoh, A. B. J. (2023). *Understanding open-set recognition by Jacobian norm and inter-class separation*. https://doi.org/10.1016/j.patcog.2023.109942

Peccoud, J., Gallegos, J., Murch, R. S., Buchholz, W. G., & Raman, S. (2017). Cyberbiosecurity: From Naive Trust to Risk Awareness. *Trends in Biotechnology*, *36*(1), 4–7. https://doi.org/10.1016/j.tibtech.2017.10.012

Porrelli, S., Fornasiero, A., Le, H. P., Yin, W., Rodriguez, M. N., Mohammed, N., Himmelbach, A., Clarke, A., Stein, N., Kersey, P., Wing, R., & Gutaker, R. M. (2026). Patterns of aDNA Damage Through Time and Environments – lessons from herbarium specimens. *GigaScience*. https://doi.org/10.1093/gigascience/giag026

Quince, C., Walker, A. W., Simpson, J. T., Loman, N. J., & Segata, N. (2017). *Shotgun metagenomics, from sampling to analysis*. https://doi.org/10.1038/nbt.3935

Rang, F. J., Kloosterman, W. P., & Ridder, J. de. (2018). *From squiggle to basepair: computational approaches for improving nanopore sequencing read accuracy*. https://doi.org/10.1186/s13059-018-1462-9

Schmidt, C. W., Reddy, V., Zhang, H., Alameddine, A., Uzan, O., Pinter, Y., & Tanner, C. C. (2024). Tokenization Is More Than Compression. In *arXiv (Cornell University)*. Cornell University. https://doi.org/10.48550/arxiv.2402.18376

Schubert, M., Ginolhac, A., Lindgreen, S., Thompson, J. F., Al-Rasheid, K. A. S., Willerslev, E., Krogh, A., & Orlando, L. (2012). Improving ancient DNA read mapping against modern reference genomes. *BMC Genomics*, *13*(1), 178–178. https://doi.org/10.1186/1471-2164-13-178

Sennrich, R., Haddow, B., & Birch, A. (2016). *Neural Machine Translation of Rare Words with Subword Units*. https://doi.org/10.18653/v1/p16-1162

Shapiro, B., & Hofreiter, M. (2014). *A Paleogenomic Perspective on Evolution and Gene Function: New Insights from Ancient DNA*. https://doi.org/10.1126/science.1236573

Skoglund, P., Northoff, B. H., Shunkov, M. V., Деревянко, А. П., Pääbo, S., Krause, J., & Jakobsson, M. (2014). Separating endogenous ancient DNA from modern day contamination in a Siberian Neandertal. *Proceedings of the National Academy of Sciences*, *111*(6), 2229–2234. https://doi.org/10.1073/pnas.1318934111

Tang, Z., Mitsui, Y., Miyazaki, T., & Omachi, S. (2026). *Multi-masking strategies for self-supervised Low- and High-level text representation learning*. https://doi.org/10.1016/j.patcog.2026.113273

Wettig, A., Gao, T., Zhong, Z., & Chen, D. (2023). *Should You Mask 15% in Masked Language Modeling?* https://doi.org/10.18653/v1/2023.eacl-main.217

Wolf, T., Debut, L., Sanh, V., Chaumond, J., Delangue, C., Moi, A., Cistac, P., Rault, T., Louf, R., Funtowicz, M., Davison, J., Shleifer, S., Platen, P. von, Ma, C., Jernite, Y., Plu, J., Xu, C., Scao, T. L., Gugger, S., … Rush, A. M. (2020). *Transformers: State-of-the-Art Natural Language Processing*. 38–45. https://doi.org/10.18653/v1/2020.emnlp-demos.6

Zhang, Y., & Skolnick, J. (2004). *Scoring function for automated assessment of protein structure template quality*. https://doi.org/10.1002/prot.20264

Zhou, Z., Ji, Y., Li, W., Dutta, P., Davuluri, R. V., & Liu, H. (2023). *DNABERT-2: Efficient Foundation Model and Benchmark For Multi-Species Genome*. https://doi.org/10.48550/arxiv.2306.15006